\pdfoutput=1

\documentclass[11pt]{article}

\usepackage[]{emnlp2021}

\usepackage{times}
\usepackage{latexsym}
\usepackage{graphicx}
\usepackage{amsmath}
\usepackage{multirow}
\usepackage{amsfonts}
\usepackage[T1]{fontenc}

\usepackage[utf8]{inputenc}

\usepackage{microtype}
\usepackage[font=small,skip=0pt]{caption}
\title{Bitext Mining for Low-Resource Languages via Contrastive Learning}

\author{Weiting Tan\and Philipp Koehn\\
  Center for Language and Speech Processing\\Computer Science Department\\Johns Hopkins University \\
  \texttt{\{wtan12, phi\}@jhu.edu}
}
\begin{document}
\maketitle

\begin{abstract}
Mining high-quality bitexts for low-resource languages is challenging. This paper shows that sentence representation of language models fine-tuned with multiple negatives ranking loss, a contrastive objective, helps retrieve clean bitexts. Experiments show that parallel data mined from our approach substantially outperform the previous state-of-the-art method on low resource languages Khmer and Pashto.
\end{abstract}

\section{Introduction}
Modern neural machine translation (NMT) system's success largely depends on the amount of high quality parallel training data. ParaCrawl\footnote{\url{https://paracrawl.eu/}}, One of the popular projects to mine bitexts, crawls webpages and retrieve sentence pairs for various languages. In this paper, we improve the quality of bitexts mined from ParaCrawl for two of the low resource languages (data size smaller than 10 million), Khmer (km) and Pashto (ps). ParaCrawl mines corpus with a pipeline of four major steps:
\begin{enumerate}
    \item Website Crawling: crawl websites and collect contents of web pages.
    \item Document Alignment: from collected web pages, find contents that align with each other in different languages. Since web page has blocks of contents, this step aligns content on document level
    \item Sentence Alignment: from aligned documents, retrieve aligned sentences by finding matched sentences pairs of two languages. 
    \item Sentence Filtering: from aligned sentences, filter out noisy sentence pairs and use the rest as clean parallel data for downstream tasks such as training NMT systems. 
\end{enumerate}
\begin{figure}[t]
    \centering
    \begin{tabular}{c}
    \includegraphics[width=0.45\textwidth]{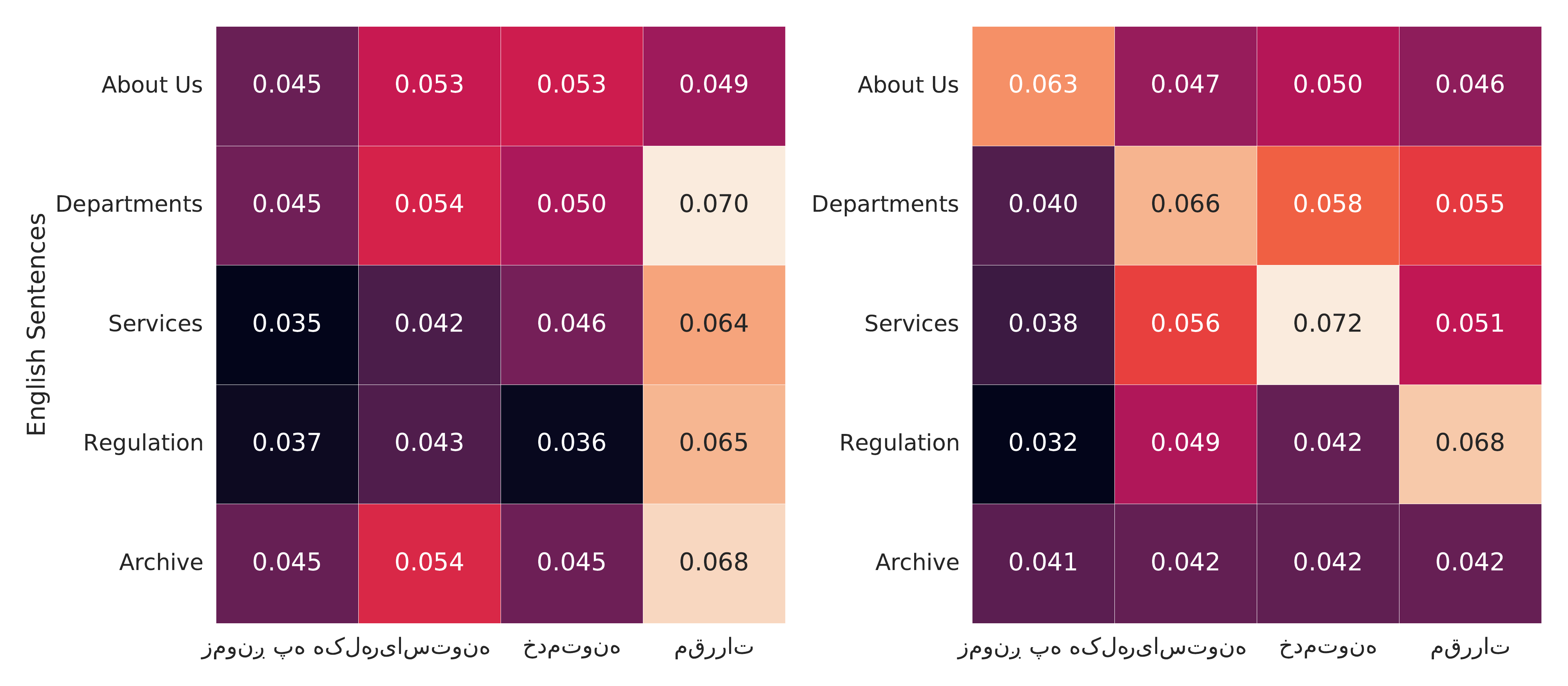} \\
    \end{tabular}
    \caption{Heatmap of cosine similarity for EN-PS sentences computed by LASER (left) and our fine-tuned Embedding (right). X-axis is Pashto sentence (left to right: About Us, Departments, Services, Regulation). Matching pairs have higher similarity score by our embedding. LASER embedding also mistakenly give "Archive" high score with "Regulation".
    \label{fig:vis}}
\end{figure}
While each step of the pipeline can be improved, we focus on sentence alignment and sentence filtering steps, both of which could benefit from an improved sentence scoring function. In this paper, we apply contrastive learning \cite{https://doi.org/10.48550/arxiv.2002.05709, mnr} to fine-tune a sentence transformer model and use it to align and filter sentences for Pashto and Khmer. Our contrastively fine-tuned sentence transformer achieves better results than previous state-of-the-art sentence representation LASER \cite{Artetxe_2019} as well as other top-performance filtering systems
\cite{acarcicek-etal-2020-filtering, lu-etal-2020-alibaba} submitted to WMT 2020 Corpus Filtering Task \cite{koehn-etal-2020-findings}. We release a toolkit\footnote{Code available at: \url{https://github.com/steventan0110/align-filter}} for replicating our experiments and it can be applied to other language pairs with document-aligned data.
\begin{figure*}[h]
    \centering
    \begin{tabular}{c}
    \includegraphics[width=0.9\textwidth]{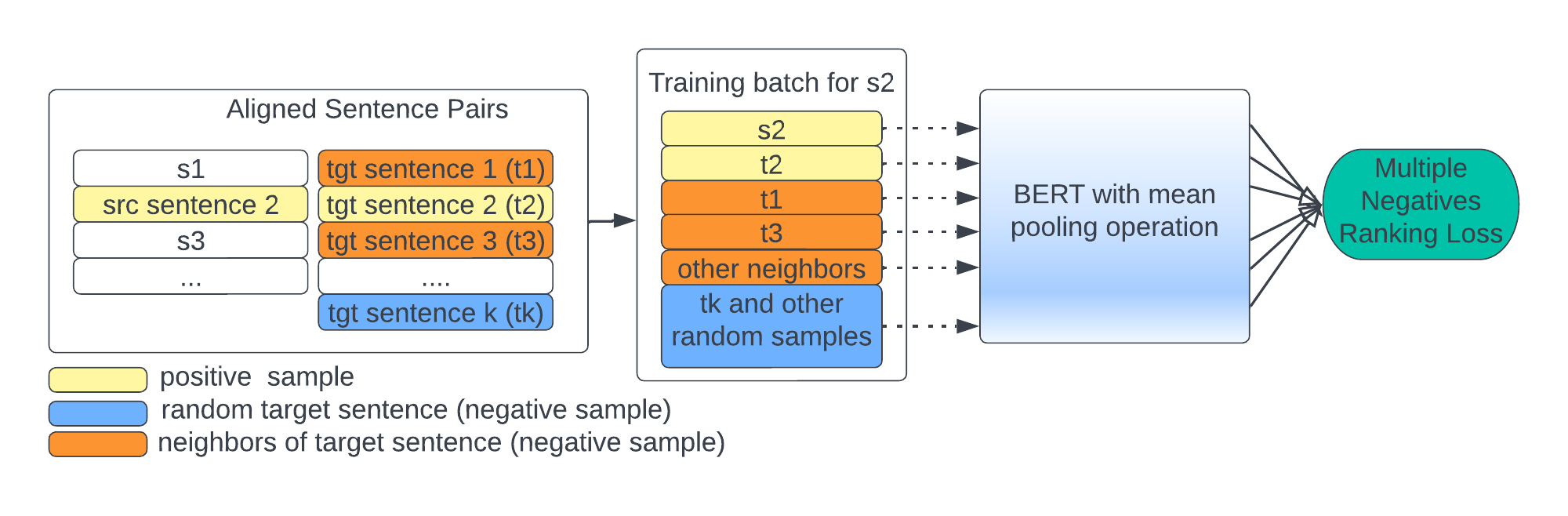} \\
    \end{tabular}
    \caption{Fine-tune Sentence Transformer (BERT) with Multiple Negatives Ranking Loss
    \label{fig:method}}
\end{figure*}
\section{Related Work}
\subsection{Sentence Alignment}
Earlier sentence aligners use heuristics such as sentence length and word frequency to find matching sentence pairs in two documents. One such tool, Hunalign \cite{Varga2007ParallelCF}, is still used today to align sentences from document pairs in the ParaCrawl project. More recent aligners such as Bleualign \cite{sennrich-volk-2010-mt} use translation system to get both documents into the same language and find matching sentence pairs. Another recent aligner, Vecalign\footnote{\url{https://github.com/thompsonb/vecalign}} \cite{thompson-koehn-2019-vecalign} uses time series warping \cite{dtw} to improve the algorithm run-time for alignment and can be used with any cross-lingual sentence embedding such as LASER to compute a similarity score.

\subsection{Sentence Pair Filtering}
Different filtering methods have been proposed in the past few years, for instance in the context of the shared tasks organized by WMT \cite{buck-koehn-2016-findings,koehn-etal-2018-findings,koehn-etal-2019-findings, koehn-etal-2020-findings} and BUCC  \cite{zweigenbaum-etal-2017-overview, zweigenbaum:hal-01898360}. The cosine similarity score computed by LASER is widely used to filter sentences because (1) pre-trained LASER embedding models are publicly available and (2) computing cosine similarity based on embeddings is fast. Another high performance method is dual conditional cross-entropy \cite{https://doi.org/10.48550/arxiv.1809.00197}, which uses translation system to find maximal symmetric agreement among sentence pairs. With recent advance on pre-trained language models such as BERT \cite{bert}, Roberta \cite{roberta} and XLM-Roberta \cite{xlm-roberta}, proxy learning (sentence filtering as a binary classification task) \cite{acarcicek-etal-2020-filtering} also performs very well. 

\section{Methodology}
\subsection{Fine-tune Sentence Transformer}\label{sec::method}
\citet{acarcicek-etal-2020-filtering} proposed proxy learning for the sentence filtering task. They used a pre-trained language model with a binary classification head to detect high quality sentence pairs. Inspired by proxy learning, sentence transformers \cite{reimers-gurevych-2019-sentence}, and contrastive learning \cite{https://doi.org/10.48550/arxiv.2002.05709}, we fine-tuned sentence transformers following figure~\ref{fig:method} to learn a sentence embedding for source and target languages. To fine-tune sentence transformers (we use sentence-BERT, or SBERT), we construct positive and negative pairs to train models with Multiple Negative Ranking Loss \cite[MNR;][]{mnr}. Given N aligned sentence pairs $\{(s_1, t_1), \cdots (s_N, t_N)\}$, each aligned pair is a positive sample. To construct negative samples, for any given source sentence $s_i$, we use window size W to take neighbors of $t_i$ to form negative samples $\{(s_i, t_{i-w}), (s_i, t_{i-w+1}), \cdots, (s_i, t_{i+w})\}$. We also randomly sample R sentences from target side to form negative samples with $s_i$. Following MNR Loss, the training objective is computed as
\begin{equation}
    \mathcal{J}(s,t, \theta)= -\frac{1}{K}\sum_{i=1}^{K}\left[d(s_i,t_i) - \log\sum_{j=1}^{2W+R} e^{d(s_i, t_j)} \right]
\end{equation}
Where K is the batch size, $(s_i, t_i)$ is the aligned source-target sentence pair (positive sample) and $(s_i, t_j)$ is the negative sample. The distance or similarity score is measured by cosine similarity: 
\begin{equation}
    d(s_i, t_i) = \cos(r_{s_i}, r_{t_i})
\end{equation}
where $r_{s_i}$ is the high dimensional sentence representation of $s_i$ encoded by the pre-trained language model $\theta$. By minimizing $\mathcal{J}(s,t,\theta)$, the model learns to maximize the difference between similarity scores of positive and negative pairs. Therefore, models fine-tuned with MNR not only recognize similar sentences but can also discard noisy sentences. The advantage of our fine-tuned models against other contrastively trained systems like \citet{acarcicek-etal-2020-filtering} is that our representation can be quickly computed for millions of sentences and then used for alignment or filtering tasks.

\subsection{Sentence Alignment}
The task of sentence alignment is to find matching sentence pairs in each aligned document pair $\{D^{\text{src}}, D^{\text{tgt}}\} = \{\{s_1, s_2\cdots,s_m, \}, \{t_1,t_2\cdots,s_n, \}\}$. We hope to retrieve k sentence pairs $\{(s_{i_1}, t_{j_1}), \cdots, (s_{i_k}, t_{i_k})\}$, where each index $i_k (j_k)$ correspond to a set of indexes in $D^{\text{src}} (D^{\text{tgt}})$. For example, $i_1 = (1,2,3), j_1 = (1)$ stands for aligning $\{s_1, s_2, s_3\}$ from source to $\{t_1\}$ from target. We use Vecalign as the alignment algorithm because it is designed to work with any high dimensional sentence embedding and it uses approximate dynamic programming algorithm that run in $O(NM)$ time  ($N$ and $M$ are the number of sentences in source and target document). Therefore we use Vecalign to quickly align sentences in document-aligned corpus by feeding in LASER or our fine-tuned SBERT embedding. For details of Vecalign, we direct readers to the original paper \footnote{\url{https://aclanthology.org/D19-1136.pdf}}. 

\subsection{Sentence Filtering}\label{sec::filter}
We replicated the filtering system from HUAWEI \cite{acarcicek-etal-2020-filtering} which rank $1^{st}$ on corpus filtering task for Pashto and $2^{nd}$ for Khmer in WMT 2020. HUAWEI's system directly fine-tune language models with a binary classification head\footnote{We follow their practice to fine-tune XLM-Roberta model} so we can filter the corpus by ranking scores predicted by the model. We also experimented with sentence representation (LASER and our fine-tuned SBERT) to filter corpus. Since we need to compute a similarity score based on two high dimensional vectors, we resort to margin score \cite{marginscore}, a similarity function that is shown to alleviate the "hubness" problem \cite{JMLR:v11:radovanovic10a, lazaridou-etal-2015-hubness}. For each given sentence pair $(x,y)$ and the encoded representation $(r_x, r_y)$, the score is computed as
\begin{equation}
\begin{split}
    &d(r_x, r_y) = \text{margin}(\cos(r_x, r_y),\\ 
    &\sum_{z\in \text{NN}_k(x)}\frac{\cos(r_x, r_z)}{2k} + \sum_{z\in \text{NN}_k (y)}\frac{\cos(r_y, r_z)}{2k})
\end{split}    
\end{equation}
where $\text{NN}_k(x)$ is the k-nearest neighbor of x in the corpus\footnote{We use Faiss \cite{johnson2019billion} to compute the score and use its default value k=4}. In practice, we use ratio as the margin function, namely $\text{margin}(a,b) = \frac{a}{b}$.

\section{Experiments and Results}
\subsection{Mined Datasets Description}
We use the evaluation setup of the WMT 2020 shared task on parallel corpus filtering\footnote{\url{https://www.statmt.org/wmt20/parallel-corpus-filtering.html}} \cite{koehn-etal-2020-findings}. Alignment and filtering methods are evaluated by training MT systems on the resulting parallel corpora and assessing their quality with BLEU \cite{papineni-etal-2002-bleu}. We start with the sentence-aligned corpus and the document-aligned corpus provided by WMT. We denote the sentence-aligned corpus as \texttt{HUNALIGN} since it is aligned by Hunalign\footnote{\url{http://mokk.bme.hu/en/resources/hunalign/}} tool. We use Vecalign with LASER embeddings and our fine-tuned SBERT embeddings on the released document-aligned corpus, producing two more versions of sentence-aligned corpora: \texttt{LASER-ALIGN} and \texttt{SBERT-ALIGN}. 



For each of the three versions of parallel corpus (\texttt{LASER-ALIGN}, \texttt{SBERT-ALIGN}, and \texttt{HUNALIGN}), we de-noise it with three filtering methods: \texttt{LASER-FILTER}, \texttt{SBERT-FILTER}, \texttt{HUAWEI-FILTER}. Both \texttt{LASER-FILTER} and \texttt{SBERT-FILTER} rank sentences with margin score function, with the only difference being the sentence representation. \texttt{HUAWEI-FILTER} is our replication of the filtering system from HUAWEI as described in section \ref{sec::filter}. 

\begin{figure*}[h]
    \centering
    \begin{tabular}{cc}
    \includegraphics[width=0.5\textwidth]{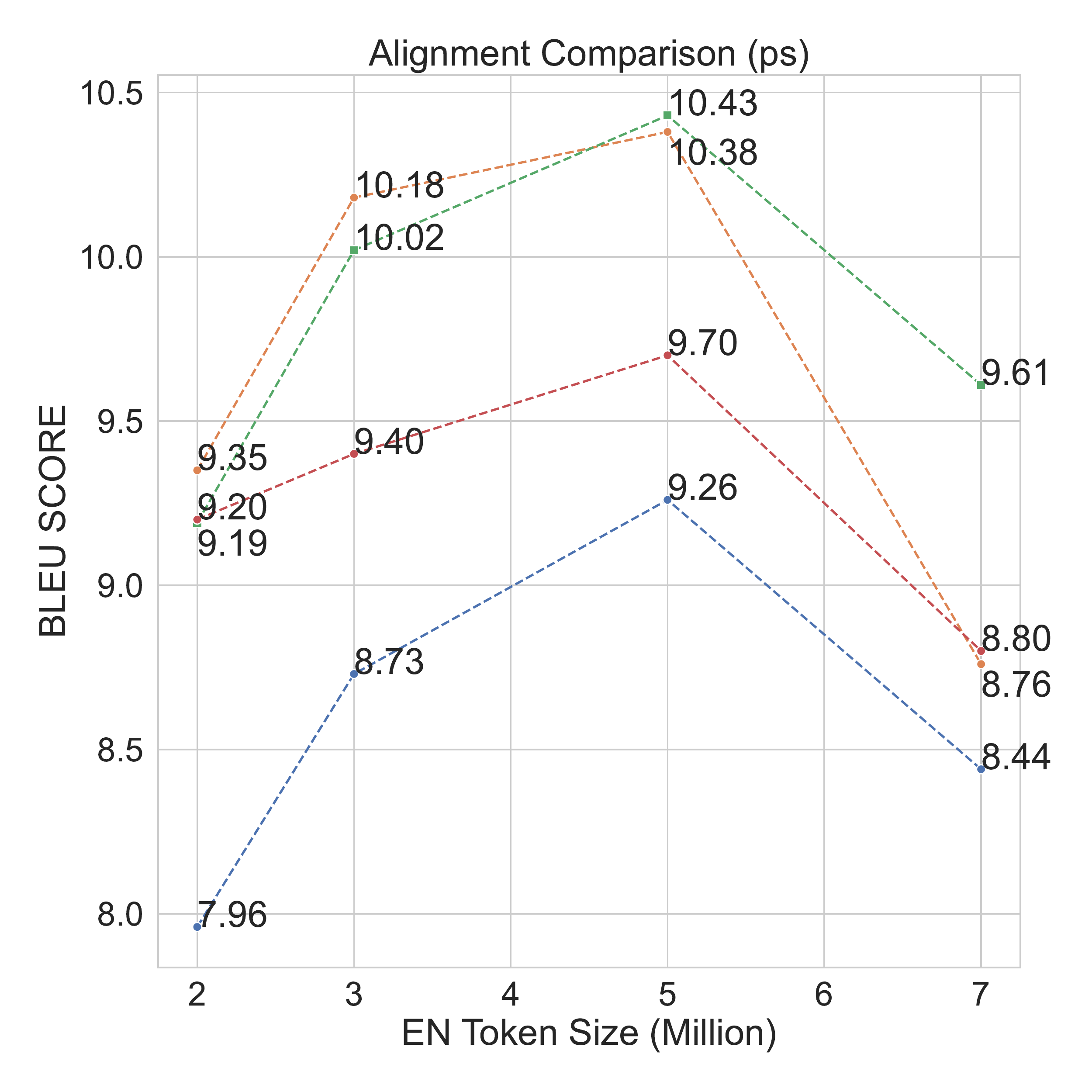} \includegraphics[width=0.5\textwidth]{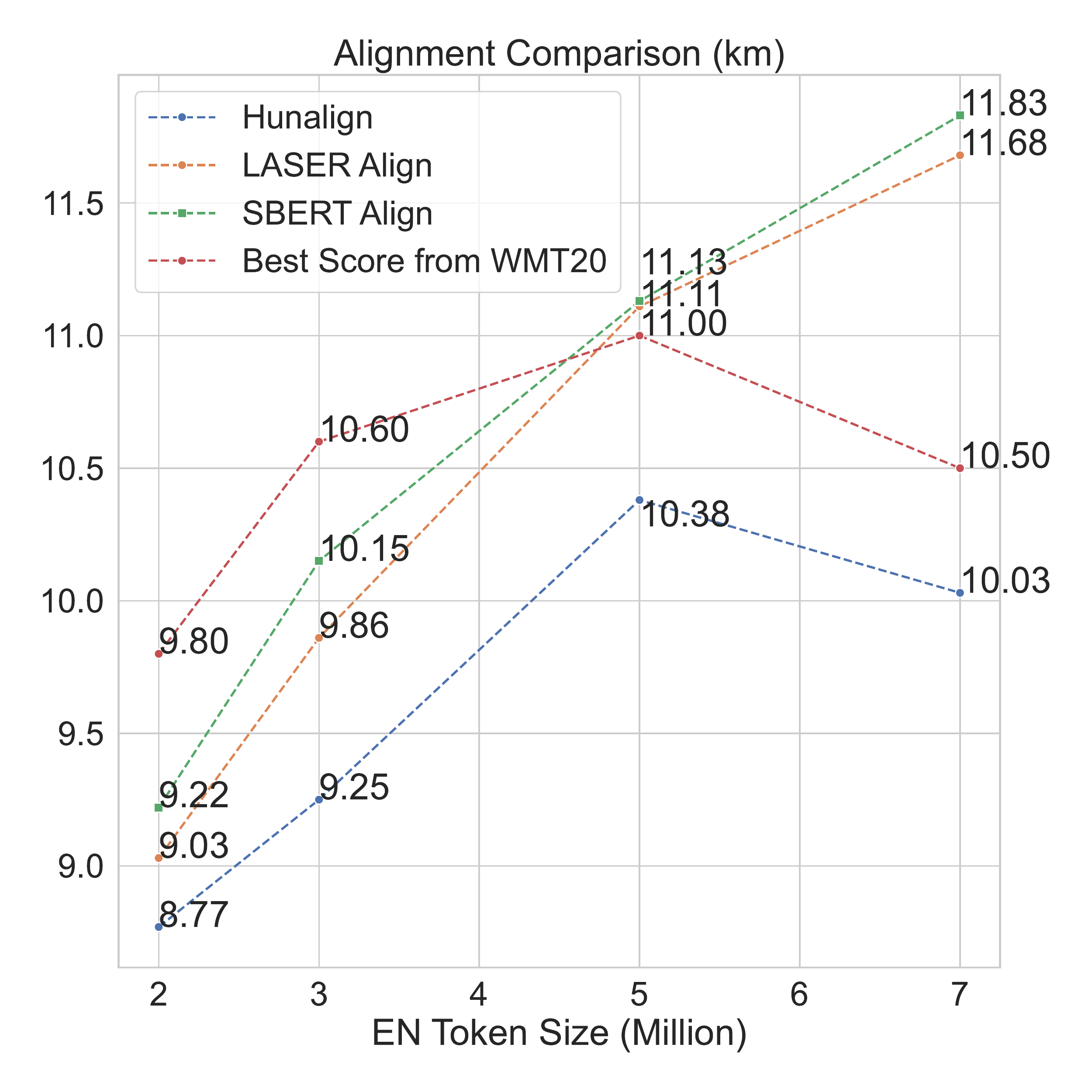}
    \end{tabular}
    \caption{Alignment method comparison among sentence-aligned corpus \texttt{HUNALIGN}, \texttt{LASER-ALIGN}, and \texttt{SBERT-ALIGN}. For each corpus, three filtering methods (LASER, SBERT, HUAWEI) are experimented and we plot the highest score out of three. The highest score of WMT2020 Corpus Filtering task is also shown here in red line for comparison (We direct readers to original papers submitted to WMT2020 for detailed description of their filtering methods).
    \label{fig:align}}
\end{figure*}

\subsection{Results and Analysis}
To evaluate performance of different methods, we rely on the BLEU score \cite{papineni-etal-2002-bleu} of the neural machine translation model trained following the \textsl{flores} baseline setting\footnote{\url{https://github.com/facebookresearch/flores}}. Complete experimental results (Table~\ref{table:ps} and~\ref{table:km}) and detailed description of preprocessing and fine-tuning steps are included in appendix. In this section, Figures~\ref{fig:align} and~\ref{fig:ls} are used to help visualize results. Figure~\ref{fig:align} shows the best BLEU score achieved by each sentence-aligned corpus. For each corpus, we experimented with three filtering techniques and only the highest score of the three is plotted. For both languages (ps and km), the best score is from \texttt{SBERT-ALIGN} (10.43 for Pashto and 11.83 for Khmer), which is about 1 BLEU point boost compared to the wining system in WMT20. The substantial difference between semantic-representation-based methods (LASER/SBERT-align) and heuristic-based method (Hunalign) can be easily found in the plot.
Between \texttt{LASER-ALIGN} and \texttt{SBERT-ALIGN}, the advantage of the latter seems small but here we only plot the highest score out of three filtering methods. In fact, most time \texttt{HUAWEI-FILTER} works the best (and it is the most computational-intensive filtering method of the three). Therefore, we plot figure~\ref{fig:ls} to better compare LASER and fine-tuned SBERT. We uses LASER or SBERT for both sentence alignment and filtering steps. SBERT-based alignment and filtering works much better than the LASER-based method (about +2 BLEU), demonstrating SBERT's effectiveness as alignment \& filtering technique.

Combining results above, we show that SBERT is a better representation of low resource languages, a better quality-scoring mechanism for sentence alignment and filtering. We believe there is still room for further improvement, since SBERT is only fine-tuned on target language and English but WMT-released document-aligned corpus are noisy, with boilerplate and sentences in the wrong language. A natural next step is to fine-tune SBERT with our proposed technique on much larger amount of data, covering more languages.

\begin{figure}[t]
    \centering
    \begin{tabular}{c}
    \includegraphics[width=0.45\textwidth]{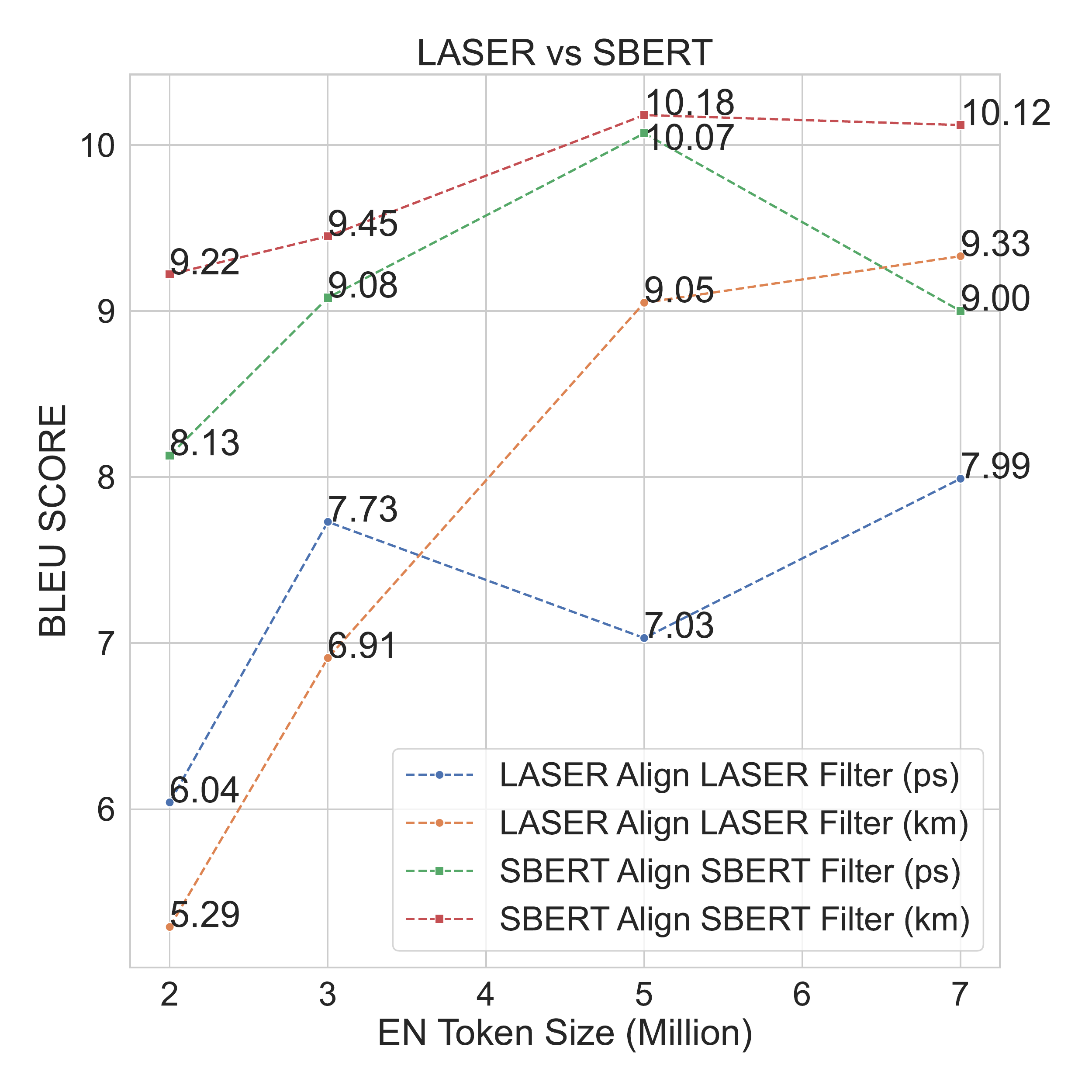}
    \end{tabular}
    \caption{LASER and fine-tuned SBERT Comparison. For both languages, using SBERT (red and green lines) to align and filter sentences results in >1 BLEU score than LASER (orange and blue lines).
    \label{fig:ls}}
\end{figure}

\section{Conclusion}
We empirically show that SBERT, fine-tuned with Multiple Negative Ranking Loss, is a good sentence representation of low resource languages. Using our fine-tuned SBERT as a sentence-aligner (with Vecalign) or filter (with margin-based score) produces better training data for downstream neural machine translation models.

\bibliography{anthology,custom}

\begin{thebibliography}{25}
\expandafter\ifx\csname natexlab\endcsname\relax\def\natexlab#1{#1}\fi

\bibitem[{A{\c{c}}ar{\c{c}}i{\c{c}}ek et~al.(2020)A{\c{c}}ar{\c{c}}i{\c{c}}ek,
  {\c{C}}olako{\u{g}}lu, Aktan~Hatipo{\u{g}}lu, Huang, and
  Peng}]{acarcicek-etal-2020-filtering}
Haluk A{\c{c}}ar{\c{c}}i{\c{c}}ek, Talha {\c{C}}olako{\u{g}}lu, P{\i}nar~Ece
  Aktan~Hatipo{\u{g}}lu, Chong~Hsuan Huang, and Wei Peng. 2020.
\newblock \href {https://aclanthology.org/2020.wmt-1.105} {Filtering noisy
  parallel corpus using transformers with proxy task learning}.
\newblock In \emph{Proceedings of the Fifth Conference on Machine Translation},
  pages 940--946, Online. Association for Computational Linguistics.

\bibitem[{Artetxe and Schwenk(2019{\natexlab{a}})}]{marginscore}
Mikel Artetxe and Holger Schwenk. 2019{\natexlab{a}}.
\newblock \href {https://doi.org/10.18653/v1/p19-1309} {Margin-based parallel
  corpus mining with multilingual sentence embeddings}.
\newblock In \emph{Proceedings of the 57th Annual Meeting of the Association
  for Computational Linguistics}. Association for Computational Linguistics.

\bibitem[{Artetxe and Schwenk(2019{\natexlab{b}})}]{Artetxe_2019}
Mikel Artetxe and Holger Schwenk. 2019{\natexlab{b}}.
\newblock \href {https://doi.org/10.1162/tacl_a_00288} {Massively multilingual
  sentence embeddings for zero-shot cross-lingual transfer and beyond}.
\newblock \emph{Transactions of the Association for Computational Linguistics},
  7:597--610.

\bibitem[{Buck and Koehn(2016)}]{buck-koehn-2016-findings}
Christian Buck and Philipp Koehn. 2016.
\newblock \href {https://doi.org/10.18653/v1/W16-2347} {Findings of the {WMT}
  2016 bilingual document alignment shared task}.
\newblock In \emph{Proceedings of the First Conference on Machine Translation:
  Volume 2, Shared Task Papers}, pages 554--563, Berlin, Germany. Association
  for Computational Linguistics.

\bibitem[{Chen et~al.(2020)Chen, Kornblith, Norouzi, and
  Hinton}]{https://doi.org/10.48550/arxiv.2002.05709}
Ting Chen, Simon Kornblith, Mohammad Norouzi, and Geoffrey Hinton. 2020.
\newblock \href {https://doi.org/10.48550/ARXIV.2002.05709} {A simple framework
  for contrastive learning of visual representations}.

\bibitem[{Conneau et~al.(2019)Conneau, Khandelwal, Goyal, Chaudhary, Wenzek,
  Guzmán, Grave, Ott, Zettlemoyer, and Stoyanov}]{xlm-roberta}
Alexis Conneau, Kartikay Khandelwal, Naman Goyal, Vishrav Chaudhary, Guillaume
  Wenzek, Francisco Guzmán, Edouard Grave, Myle Ott, Luke Zettlemoyer, and
  Veselin Stoyanov. 2019.
\newblock \href {https://doi.org/10.48550/ARXIV.1911.02116} {Unsupervised
  cross-lingual representation learning at scale}.

\bibitem[{Devlin et~al.(2018)Devlin, Chang, Lee, and Toutanova}]{bert}
Jacob Devlin, Ming-Wei Chang, Kenton Lee, and Kristina Toutanova. 2018.
\newblock \href {https://doi.org/10.48550/ARXIV.1810.04805} {Bert: Pre-training
  of deep bidirectional transformers for language understanding}.

\bibitem[{Henderson et~al.(2017)Henderson, Al-Rfou, Strope, Sung, Lukacs, Guo,
  Kumar, Miklos, and Kurzweil}]{mnr}
Matthew Henderson, Rami Al-Rfou, Brian Strope, Yun-hsuan Sung, Laszlo Lukacs,
  Ruiqi Guo, Sanjiv Kumar, Balint Miklos, and Ray Kurzweil. 2017.
\newblock \href {https://doi.org/10.48550/ARXIV.1705.00652} {Efficient natural
  language response suggestion for smart reply}.

\bibitem[{Johnson et~al.(2019)Johnson, Douze, and
  J{\'e}gou}]{johnson2019billion}
Jeff Johnson, Matthijs Douze, and Herv{\'e} J{\'e}gou. 2019.
\newblock Billion-scale similarity search with {GPUs}.
\newblock \emph{IEEE Transactions on Big Data}, 7(3):535--547.

\bibitem[{Junczys-Dowmunt(2018)}]{https://doi.org/10.48550/arxiv.1809.00197}
Marcin Junczys-Dowmunt. 2018.
\newblock \href {https://doi.org/10.48550/ARXIV.1809.00197} {Dual conditional
  cross-entropy filtering of noisy parallel corpora}.

\bibitem[{Koehn et~al.(2020)Koehn, Chaudhary, El-Kishky, Goyal, Chen, and
  Guzm{\'a}n}]{koehn-etal-2020-findings}
Philipp Koehn, Vishrav Chaudhary, Ahmed El-Kishky, Naman Goyal, Peng-Jen Chen,
  and Francisco Guzm{\'a}n. 2020.
\newblock \href {https://aclanthology.org/2020.wmt-1.78} {Findings of the {WMT}
  2020 shared task on parallel corpus filtering and alignment}.
\newblock In \emph{Proceedings of the Fifth Conference on Machine Translation},
  pages 726--742, Online. Association for Computational Linguistics.

\bibitem[{Koehn et~al.(2019)Koehn, Guzm{\'a}n, Chaudhary, and
  Pino}]{koehn-etal-2019-findings}
Philipp Koehn, Francisco Guzm{\'a}n, Vishrav Chaudhary, and Juan Pino. 2019.
\newblock \href {https://doi.org/10.18653/v1/W19-5404} {Findings of the {WMT}
  2019 shared task on parallel corpus filtering for low-resource conditions}.
\newblock In \emph{Proceedings of the Fourth Conference on Machine Translation
  (Volume 3: Shared Task Papers, Day 2)}, pages 54--72, Florence, Italy.
  Association for Computational Linguistics.

\bibitem[{Koehn et~al.(2018)Koehn, Khayrallah, Heafield, and
  Forcada}]{koehn-etal-2018-findings}
Philipp Koehn, Huda Khayrallah, Kenneth Heafield, and Mikel~L. Forcada. 2018.
\newblock \href {https://doi.org/10.18653/v1/W18-6453} {Findings of the {WMT}
  2018 shared task on parallel corpus filtering}.
\newblock In \emph{Proceedings of the Third Conference on Machine Translation:
  Shared Task Papers}, pages 726--739, Belgium, Brussels. Association for
  Computational Linguistics.

\bibitem[{Lazaridou et~al.(2015)Lazaridou, Dinu, and
  Baroni}]{lazaridou-etal-2015-hubness}
Angeliki Lazaridou, Georgiana Dinu, and Marco Baroni. 2015.
\newblock \href {https://doi.org/10.3115/v1/P15-1027} {Hubness and pollution:
  Delving into cross-space mapping for zero-shot learning}.
\newblock In \emph{Proceedings of the 53rd Annual Meeting of the Association
  for Computational Linguistics and the 7th International Joint Conference on
  Natural Language Processing (Volume 1: Long Papers)}, pages 270--280,
  Beijing, China. Association for Computational Linguistics.

\bibitem[{Liu et~al.(2019)Liu, Ott, Goyal, Du, Joshi, Chen, Levy, Lewis,
  Zettlemoyer, and Stoyanov}]{roberta}
Yinhan Liu, Myle Ott, Naman Goyal, Jingfei Du, Mandar Joshi, Danqi Chen, Omer
  Levy, Mike Lewis, Luke Zettlemoyer, and Veselin Stoyanov. 2019.
\newblock \href {https://doi.org/10.48550/ARXIV.1907.11692} {Roberta: A
  robustly optimized bert pretraining approach}.

\bibitem[{Lu et~al.(2020)Lu, Ge, Shi, and Zhang}]{lu-etal-2020-alibaba}
Jun Lu, Xin Ge, Yangbin Shi, and Yuqi Zhang. 2020.
\newblock \href {https://aclanthology.org/2020.wmt-1.111} {{A}libaba submission
  to the {WMT}20 parallel corpus filtering task}.
\newblock In \emph{Proceedings of the Fifth Conference on Machine Translation},
  pages 979--984, Online. Association for Computational Linguistics.

\bibitem[{Papineni et~al.(2002)Papineni, Roukos, Ward, and
  Zhu}]{papineni-etal-2002-bleu}
Kishore Papineni, Salim Roukos, Todd Ward, and Wei-Jing Zhu. 2002.
\newblock \href {https://doi.org/10.3115/1073083.1073135} {{B}leu: a method for
  automatic evaluation of machine translation}.
\newblock In \emph{Proceedings of the 40th Annual Meeting of the Association
  for Computational Linguistics}, pages 311--318, Philadelphia, Pennsylvania,
  USA. Association for Computational Linguistics.

\bibitem[{Radovanović et~al.(2010)Radovanović, Nanopoulos, and
  Ivanović}]{JMLR:v11:radovanovic10a}
Milo{\v{s}} Radovanović, Alexandros Nanopoulos, and Mirjana Ivanović. 2010.
\newblock \href {http://jmlr.org/papers/v11/radovanovic10a.html} {Hubs in
  space: Popular nearest neighbors in high-dimensional data}.
\newblock \emph{Journal of Machine Learning Research}, 11(86):2487--2531.

\bibitem[{Reimers and Gurevych(2019)}]{reimers-gurevych-2019-sentence}
Nils Reimers and Iryna Gurevych. 2019.
\newblock \href {https://doi.org/10.18653/v1/D19-1410} {Sentence-{BERT}:
  Sentence embeddings using {S}iamese {BERT}-networks}.
\newblock In \emph{Proceedings of the 2019 Conference on Empirical Methods in
  Natural Language Processing and the 9th International Joint Conference on
  Natural Language Processing (EMNLP-IJCNLP)}, pages 3982--3992, Hong Kong,
  China. Association for Computational Linguistics.

\bibitem[{Salvador and Chan(2007)}]{dtw}
Stan Salvador and Philip Chan. 2007.
\newblock Toward accurate dynamic time warping in linear time and space.
\newblock \emph{Intell. Data Anal.}, 11(5):561–580.

\bibitem[{Sennrich and Volk(2010)}]{sennrich-volk-2010-mt}
Rico Sennrich and Martin Volk. 2010.
\newblock \href {https://aclanthology.org/2010.amta-papers.14} {{MT}-based
  sentence alignment for {OCR}-generated parallel texts}.
\newblock In \emph{Proceedings of the 9th Conference of the Association for
  Machine Translation in the Americas: Research Papers}, Denver, Colorado, USA.
  Association for Machine Translation in the Americas.

\bibitem[{Thompson and Koehn(2019)}]{thompson-koehn-2019-vecalign}
Brian Thompson and Philipp Koehn. 2019.
\newblock \href {https://doi.org/10.18653/v1/D19-1136} {{V}ecalign: Improved
  sentence alignment in linear time and space}.
\newblock In \emph{Proceedings of the 2019 Conference on Empirical Methods in
  Natural Language Processing and the 9th International Joint Conference on
  Natural Language Processing (EMNLP-IJCNLP)}, pages 1342--1348, Hong Kong,
  China. Association for Computational Linguistics.

\bibitem[{Varga et~al.(2007)Varga, Hal{\'a}csy, Kornai, Viktor, Laszlo,
  L{\'a}szl{\'o}, and Viktor}]{Varga2007ParallelCF}
D{\'a}niel Varga, P{\'e}ter Hal{\'a}csy, Andr{\'a}s Kornai, Nagy Viktor, Nagy
  Laszlo, N{\'e}meth L{\'a}szl{\'o}, and Tron Viktor. 2007.
\newblock Parallel corpora for medium density languages.

\bibitem[{Zweigenbaum et~al.(2017)Zweigenbaum, Sharoff, and
  Rapp}]{zweigenbaum-etal-2017-overview}
Pierre Zweigenbaum, Serge Sharoff, and Reinhard Rapp. 2017.
\newblock \href {https://doi.org/10.18653/v1/W17-2512} {Overview of the second
  {BUCC} shared task: Spotting parallel sentences in comparable corpora}.
\newblock In \emph{Proceedings of the 10th Workshop on Building and Using
  Comparable Corpora}, pages 60--67, Vancouver, Canada. Association for
  Computational Linguistics.

\bibitem[{Zweigenbaum et~al.(2018)Zweigenbaum, Sharoff, and
  Rapp}]{zweigenbaum:hal-01898360}
Pierre Zweigenbaum, Serge Sharoff, and Reinhard Rapp. 2018.
\newblock \href {https://hal.archives-ouvertes.fr/hal-01898360} {{Overview of
  the Third BUCC Shared Task: Spotting Parallel Sentences in Comparable
  Corpora}}.
\newblock In \emph{{Workshop on Building and Using Comparable Corpora}},
  Miyazaki, Japan.

\end{thebibliography}
\bibliographystyle{acl_natbib}
\clearpage
\section{Appendix}
\subsection{Evaluation Dataset Size}
Our experiments' results are shown in Table \ref{table:ps} and \ref{table:km}, where three types of alignment method and three types are filtering methods are experimented (in total 9 combinations). For each of the 9 possible alignment-filtering method, 4 versions are created based on how much data is sub-sampled. We use the threshold 2, 3, 5, 7 million ($\#$tokens on English side) following the practice from WMT 2020 Corpus Filtering Task. Note that for Khmer, we see that the BLEU scores are still going up for some alignment-filtering methods (for example, for \texttt{SBERT-ALIGN HUAWEI-FILTER}, BLEU goes up from 11.13 to 11.83). Therefore we also experimented with sub-sampling 9-million datasets and verified that BLEU score did not increase anymore.

\subsection{Preprocessing}
For sentence alignment step, we did not employ any pre-processing techniques because most documents contain noisy sentences and removing those sentences would make it harder to align sentences. After retrieving sentence-aligned datasets (\texttt{LASER-ALIGN}, \texttt{SBERT-ALIGN}, \texttt{HUNALIGN}), we pre-process the datasets before sentence filtering step. First, we de-duplicate the datasets, which filters out about 90\% data (since most aligned sentences are duplicate pairs). Second, we remove the sentence pair that has over 90\% overlap between source and target side sentences. Lastly, We also use fasttext language id\footnote{\url{https://fasttext.cc/docs/en/language-identification.html}} to check every aligned sentence pair and remove it if its English side is not predicted as en. Note that this is a very lenient filter given the noisy sentence-aligned dataset we retrieved. In fact, language id filtering plays an important role for \texttt{LASER-FILTER} for km-en task. The BLEU score under \texttt{LASER-FILTER} is significantly worse than the other two filtering methods, especially when sub-sample size is small (2 or 3 million tokens). This is because LASER would select many sentence pairs that is not Khmer as top-scoring pairs. When filtering out sentences based on language id for both English and Khmer, \texttt{LASER-FILTER} can achieve better performances (though still worse than our \texttt{SBERT-FILTER} and \texttt{HUAWEI-FILTER} results), similar to the scores reported in WMT 2020 Corpus Filtering Task.
\subsection{Fine-tune Sentence-BERT}
To build \texttt{SBERT-ALIGN} corpus, we fine-tune the SBERT model as described in section~\ref{sec::method} and figure~\ref{fig:method}. To fine-tune SBERT, we need a parallel corpus to sample positive and negative pairs from. We experimented with both \texttt{HUNALIGN} and \texttt{LASER-ALIGN} corpus. It is unsurprising that \texttt{LASER-ALIGN} works better because it has more correctly aligned sentences. Thus, we fine-tuned SBERT based on the \texttt{LASER-ALIGN} corpus and then use it to align sentences from document-aligned data, producing the sentence-aligned corpus \texttt{SBERT-ALIGN}.

\begin{table*}
  \centering
  \renewcommand{\arraystretch}{1.2}
  \begin{tabular}{|p{3cm}|c|c|c|c|c|c|c|c|c|c|c|c|}
    \hline
    \multirow{2}{3cm}{\textbf{Alignment Type}} & \multicolumn{4}{c|}{\textbf{LASER Filter}} & 
    \multicolumn{4}{c|}{\textbf{SBERT Filter}} &
    \multicolumn{4}{c|}{\textbf{HUAWEI Filter}}\\
    \cline{2-13}
    & \textbf{2M} &\textbf{3M} & \textbf{5M} &\textbf{7M}
    & \textbf{2M} &\textbf{3M} & \textbf{5M} &\textbf{7M}
    & \textbf{2M} &\textbf{3M} & \textbf{5M} &\textbf{7M}\\
    \hline
    HUNALIGN & 6.19 & 7.53 & 7.71 & 7.86 & \textbf{7.96} & 8.73 & 9.05 & \textbf{8.44} & 7.48 & \textbf{8.70} & \textbf{9.26} & 8.33\\ \hline
    LASER-ALIGN & 6.04 & 7.73 & 7.03 & 7.99 & 8.41 & 9.17 & 9.82 & \textbf{8.76} & \textbf{9.35} & \textbf{10.18} & \textbf{10.38} & 8.70\\ \hline
    SBERT-ALIGN & 6.13& 7.72 & 8.66 & 8.44 & 8.13 & 9.08 & 10.07 & 9.00 & \textbf{9.19} & \textbf{10.02} & \textbf{10.43} & \textbf{9.61}\\ \hline
  \end{tabular}
  \caption{BLEU Score of Neural Machine Translation model trained on \textbf{ps-en} datasets that are mined with different sentence-alignment and sentence-filtering methods. Each row corresponds to one sentence-aligner and each column corresponds to one sentence-filter. Under each filter methods, four sub-columns indicate the number of millions of English tokens sub-sampled from the corpus following the practice from WMT20 Shared Task on Corpus Filtering \cite{acarcicek-etal-2020-filtering}. The bold numbers are the highest scores for each alignment type, which are used to plot figure \ref{fig:align} }
  \label{table:ps}
\end{table*}

\begin{table*}
  \centering
  \renewcommand{\arraystretch}{1.2}
  \begin{tabular}{|p{3cm}|c|c|c|c|c|c|c|c|c|c|c|c|}
    \hline
    \multirow{2}{3cm}{\textbf{Alignment Type}} & \multicolumn{4}{c|}{\textbf{LASER Filter}} & 
    \multicolumn{4}{c|}{\textbf{SBERT Filter}} &
    \multicolumn{4}{c|}{\textbf{HUAWEI Filter}}\\
    \cline{2-13}
    & \textbf{2M} &\textbf{3M} & \textbf{5M} &\textbf{7M}
    & \textbf{2M} &\textbf{3M} & \textbf{5M} &\textbf{7M}
    & \textbf{2M} &\textbf{3M} & \textbf{5M} &\textbf{7M}\\
    \hline
    HUNALIGN & 5.60 & 6.70 & 7.59 & 8.37 & 7.94 &8.27 &8.33 & 8.02 & \textbf{8.77} & \textbf{9.25} & \textbf{10.38} & \textbf{10.03}\\ \hline
    LASER-ALIGN & 5.29 & 6.91 & 9.05 & 9.33 & \textbf{9.03} & 9.44 & 10.13 & 10.79 & 8.96 & \textbf{9.86} & \textbf{11.11} & \textbf{11.68}\\ \hline
    SBERT-ALIGN & 4.96 & 7.02 & 8.65 & 9.14 & \textbf{9.22} & 9.45 & 10.18 & 10.12 & 8.85 & \textbf{10.15} & \textbf{11.13} & \textbf{11.83}\\ \hline
  \end{tabular}
  \caption{BLEU Score of Neural Machine Translation model trained on \textbf{km-en} datasets that are mined with different sentence-alignment and sentence-filtering methods. The bold numbers are the highest scores for each alignment type, which are used to plot figure \ref{fig:align}}
  \label{table:km}
\end{table*}

\end{document}